# HC-Ref: Hierarchical Constrained Refinement for Robust Adversarial Training of GNNs

Xiaobing Pei, Haoran Yang, and Gang Shen

*Abstract*—Graph Neural Networks (GNNs) apply deep neural networks to graph data and have achieved excellent performance in many tasks, such as node classification. However, recent studies have shown that attackers can catastrophically reduce the performance of GNNs by maliciously modifying the graph structure or node features on the graph. Adversarial training, which has been shown to be one of the most effective defense mechanisms against adversarial attacks in computer vision, holds great promise for enhancing the robustness of GNNs. There is limited research on defending against attacks by performing adversarial training on graphs, and it is crucial to delve deeper into this approach to optimize its effectiveness. Therefore, based on robust adversarial training on graphs, we propose a hierarchical constraint refinement framework (HC-Ref) that enhances the anti-perturbation capabilities of GNNs and downstream classifiers separately, ultimately leading to improved robustness. We propose corresponding adversarial regularization terms that are conducive to adaptively narrowing the domain gap between the normal part (clean samples) and the perturbation part (adversarial samples) according to the characteristics of different layers, promoting the smoothness of the predicted distribution of both parts. Moreover, existing research on graph robust adversarial training primarily concentrates on training from the standpoint of node feature perturbations and seldom takes into account alterations in the graph structure. This limitation makes it challenging to prevent attacks based on topological changes in the graph. This paper generates adversarial examples by utilizing graph structure perturbations, offering an effective approach to defend against attack methods that are based on topological changes. Extensive experiments on two real-world graph benchmarks show that HC-Ref successfully resists various attacks and has better node classification performance compared to several baseline methods.

This paragraph of the first footnote will contain the date on which you submitted your paper for review, which is populated by IEEE. It is IEEE style to display support information, including sponsor and financial support acknowledgment, here and not in an acknowledgment section at the end of the article. For example, "This work was supported in part by the U.S. Department of Commerce under Grant 123456." The name of the corresponding author appears after the financial information, e.g. *(Corresponding author: Second B. Author)*. Here you may also indicate if authors contributed equally or if there are co-first authors.

X. Pei is with the School of Software, Huazhong University of Science and Technology, Wuhan 430074, China (email: xiaobingp@hust.edu.cn).

H. Yang is with the School of Software, Huazhong University of Science and Technology, Wuhan 430074, China (email: m202276630@hust.edu.cn).

G. Shen is with the School of Software, Huazhong University of Science and Technology, Wuhan 430074, China (email: Gang_shen@126.com).

Color versions of one or more of the figures in this article are available online at http://ieeexplore.ieee.org

## I. INTRODUCTION

GRAPH is a ubiquitous data structure that can efficiently represent various objects and their complex relationships. This type of representation, which captures complex relationships, can be applied to model diverse datasets in multiple fields, including pharmaceutical molecules, social networks, and product recommendations. GNNs [1]–[3] acquire graph representations by leveraging the features of nodes and the topology of the graph, exhibiting exceptional performance on graph-structured data. The success of GNNs hinges on their distinctive neural information transmission mechanism, which considers node features and hidden representations as node information and propagates them through edges. However, despite the remarkable success in graph data inferring, the inherent challenge of lacking adversarial robustness in deep learning models still brings a security risk.

Recent studies have revealed that GNNs are susceptible to adversarial attacks [4]–[7]. By adding or deleting just a few edges or modifying some of the features of certain nodes, attackers can easily change the model's predictions and catastrophically decrease the performance of GNNs. To alleviate the potential consequences of these risks, enhancing model robustness is essential in some security-related application areas, such as anomaly detection [8] and fraud detection [9]. Most existing attack algorithms [10]–[13] focus on altering the graph structure to undermine the model, as some studies [5], [14] have found that changes in graph structure have a greater impact on prediction results than alterations in node features. One intuitive explanation is that adding edges between nodes with significant differences can result in the propagation of misleading information, whereas removing edges between similar nodes can hinder the dissemination of accurate information. This suggests that it is imperative to implement defense strategies from the perspective of graph structure perturbation.

One prominent approach for resisting attacks is adversarial training [15]. It employs adversarial samples rather than natural samples for training. In adversarial training, the perturbation direction is determined by calculating the loss gradient, and the perturbation is then added to the original features to create adversarial samples. Intuitively, adversarial training is chiefly a form of game training aimed at maximizing perturbation while minimizing the adversarial expected risk, thereby enabling models to withstand worst-case perturbations on input features. Numerous related works have demonstrated the impressive anti-perturbation ability of models after adversarial training [15]–[19], which implies that adversarial training holds significant promise as a means of bolstering the robustness of GNNs.

Despite the existence of some techniques that have been devised for exploring adversarial training on graphs [20]–[23], the bulk of research in this domain has concentrated on manipulating node features. To be more specific, some adversarial examples formed by adding perturbations to the features of certain nodes are designed for training robust models. For example, BVAT [22] introduced virtual adversarial training (VAT) [24] in Adversarial Training for Graph Convolutional Networks, constructing adversarial examples based on the loss gradient direction of labeled nodes and the local distributional smoothness (LDS) of unlabeled nodes. Nevertheless, as a result of insufficient specialized design and examination, there is no evidence to indicate that these models possess defensive capabilities against attacks targeting the graph structure.

To our knowledge, there are few existing works on adversarial defense training based on graph structure perturbation. Dai et al. [4] were the first to introduce structural perturbations into graph adversarial training by generating adversarial samples in a resource-efficient manner that involved randomly dropping edges globally. However, due to the excessively simplistic methodology, it has had minimal influence on improving robustness. Additionally, because of the introduction of randomness, its impact on robustness is not stable. GraphAT [25] considers the impact of the edges and neighbors, but fundamentally still perturbs node features based on a fixed graph topology. In that case, structural changes in the graph are bound to significantly decrease the model's performance. DefNet [26] introduced a conditional GAN to address the problem of generating adversarial samples in the discrete space of graph structure, and uses adversarial contrastive learning to train the GNN. SAT [27] generates graph structure perturbations using FGA [28] and NETTACK [5], and trains a robust model with smooth distillation strategy on the adversarial graph. TGD [29] introduced the concept of PGD [15] into the attack and defense models of GNNs, for the first time generating perturbed graph topology from an optimization perspective and carrying out iterative robust adversarial training. However, the majority of existing methods only train on distorted graph structures. We analyzed the inconsistency between generally using the empirical risk minimization (ERM) and training a more robust model under this setting. Adding the original structure as a regularization term into the loss to guide the training process can help propagate correct information, push the decision boundary of the model away from the sample instances, and establish a more robust model. In addition, a common way to train deep neural networks is one-step learning from representation learning to downstream tasks, directly updating the parameters of the entire model through the gradient of the loss of target task. In contrast to the aforementioned, in adversarial training, some research focuses on refining certain layers in the network, including freezing the classifier during adaptation and focusing on improving feature extractor [32], or adversarial training on penultimate activations [31]. By incorporating these training strategies, the knowledge extraction and disturbance resistance abilities of specific layers are strengthened, leading to enhancements in robustness. Such approaches encourage a focused approach towards refining the model from a layer-specific perspective.

Given the above analysis, we posit that adversarial training can lead to the development of robust models that are capable of effectively defending against graph structure attacks. In such adversarial scenarios, two key factors should be emphasized: 1) incorporating the original graph and adversarial graph together to create a regularization term within the training framework, rather than solely utilizing the adversarial graph, can aid in the establishment of a more robust decision boundary; 2) adopting a hierarchical approach to refine the model involves designing distinct tasks for separately training the feature extractor and downstream classifier.

To deal with the above challenges, in this paper, we propose a novel method called HC-Ref, namely Hierarchical Constrained Refinement for Robust Adversarial Training of GNNs. Specifically, our approach utilizes the method of convex relaxation and first-order attacks [29] to generate adversarial examples from the perspective of structural perturbations. Then, the adversarial examples are used for robust training. Different from the previous methods [4], [26], [27], [29] that solely train on the perturbed graph, we incorporate the guidance signal of the original structure into the adversarial training process. To promote the smoothness of node embedding and predicted distributions for both perturbed and original graphs, we implement dynamic regularizations, which encourages the model to establish a more robust decision boundary. Meanwhile, different task objectives are designed to train feature extractors and downstream classifiers in different phases, hierarchically improving the knowledge extraction and inference capabilities of the model. Our experimental results demonstrate that this approach is highly effective in improving adversarial robustness. In addition, in the semi-supervised scenario of adversarial training on graphs, where a low label rate is encountered, we utilize the Robust Self-Training strategy [36] to train a more robust model, which proves to be more effective than solely using a small amount of real labels. Finally, a large number of experiments show that HC-Ref can effectively defend against various adversarial attacks and achieves superior classification performance compared to some of the most advanced methods currently available.

Our contribution can be summarized as follows:

1) We propose a novel method HC-Ref for graph adversarial defense. In topology-based GNN adversarial training, a smoothness regularization term is introduced. The incorporation of the regularization containing the original structure can reduce the model's sensitivity to topological cha-nges, enhance its generalization ability, and defend against structural attacks by means of smoothing.

TABLE I
DESCRIPTION OF SOME MAJOR NOTATIONS

| Notation | Description |
| --- | --- |
| $G, G^*$ | The original graph and the perturbed graph |
| $A, A^*$ | The adjacency matrix and the perturbed adjacency matrix |
| $S, \hat{S}$ | The continuous topological perturbation matrix and the binary topological perturbation matrix |
| $V, v_i$ | The set of node and the node $i$ |
| $X, x_i$ | The feature matrix of all nodes and the feature vertor of node $v_i$ |
| $E, e_{ij}$ | The edge set of input graph and the edge $(i, j)$ |
| $D$ | The diagonal degree matrix |
| $d$ | The number of node features |
| $M$ | The number of labeled nodes |
| $N$ | The number of all nodes |
| $f(X, A\|W)$ | The output logits with model parameters $W$ when the input are $X$ and $A$ |
| $\hat{f}(X, A\|W)$ | The predited labels with model parameters $W$ when the input are $X$ and $A$ |
| $H, h_i$ | The feature matrix of hidden layer and the hidden feature vector of node $v_i$ |
| $Y, y_i$ | The real labels and the label of node $v_i$ |
| $\hat{Y}, \hat{y}_i$ | The predicted labels and the predicted label of node $v_i$ |
| $L_{p_i}$ | The objective function of phase $i$ |
| $\alpha$ | the trading hyper-parameter for adversarial regularizers in refining feature extractor |
| $\beta$ | the trading hyper-parameter for adversarial regularizers in refining classifier |

2) A new training strategy, Hierarchical Refinement, is proposed in the adversarial training scenario. Different goals are designed for the feature extractor and classifier, respectively, corresponding to the reinforcement of knowledge extraction ability and inferring ability of the model.

3) Thorough experiments conducted on two real-world datasets demonstrate that HC-Ref is highly effective in defending against various adversarial attacks and outperforms some of the most advanced methods. Additionally, we have also designed experiments to explore the reasons behind HC-Ref 's superior performance and other possible improvements.

The remainder of this article is arranged as follows. Section II overviews the preliminaries related to graph neural networks and adversarial training on graphs. Section III delineates the HC-Ref approach and highlights the benefits of the proposed method. In Section IV, the effectiveness and superiority of the HC-Ref are verified through experiments. Section V discusses related work, followed by the conclusion and the prospect of the future work in Section VI.

## II. PRELIMINARIES

Before introducing our method, we provide the necessary definitions and formulate the problem in this section. The major notations used in this paper can be found in Table I.

### A. Graph Neural Networks

Prior to defining GNN, we first introduce the following graph notations. Formally, Let $G = \{V, E, X\}$ denote an undirected and unweighted graph, where $V = \{v_1, ..., v_n\}$ denotes the set of nodes in the graph, $E = \{e_{ij} | v_i, v_j \in V\}$ is the edge set, and $X \in \mathbb{R}^{N \times d}$ represents the feature matrix. Let $A \in \{0,1\}^{N \times N}$ represent the adjacency matrix of $G$, in which $A_{ij} \in \{0,1\}$ denotes the existence of the edge $e_{ij} \in E$ that links node $v_i$ and $v_j$. We assume that each node $i$ is associated with a feature vector $x_i \in \mathbb{R}^d$ and a scalar label $y_i$. The goal of GNN is to predict the class of an unlabeled node under the graph topology $A$ and the training data $\{(x_i, y_i)\}_{i=1}^{N_{train}}$. In practice, the features of all nodes and the labels of $N_{train} < N$ nodes are used in the training phase.

GNN can be generally specified as $\hat{f}(X, A|W)$, where $W$ is the trainable parameter. Each layer of GNN can be divided into a message passing function $m_i^k = MSP(\{h_j^{k-1}; j \in \mathbb{N}_i\})$ and an updating function $h_i^k = UPD(m_i^k, h_i^{k-1})$, where $\mathbb{N}_i$ denotes the neighborhood of node $v_i$, $h_j^{k-1}$ denotes the hidden representation in the previous layer, and $h_j^0 = x_j$. The former aggregates information from neighbor nodes, and then the latter updates the representation of nodes, which is usually a sum or concat function. Though a number of different GNN methods have been proposed, here we focus on a representative one, that is, graph convolutional networks (GCN) [1]. The GCN model follows the following rule to aggregate the neighboring features:

$$H^{(k)} = \sigma(\widetilde{D}^{-\frac{1}{2}} \widetilde{A} \widetilde{D}^{-\frac{1}{2}} H^{(k-1)} W^{(k-1)}) \quad (1)$$

where $\widetilde{A} = A + I_N$ is the adjacency matrix of the graph $G$ with self connections added, $\widetilde{D}$ is a diagonal matrix with $\widetilde{D}_{i,i} = \sum_j \widetilde{A}_{ij}$ and $\sigma$ is the activation function that introduces non-linearity. Each equation above corresponds to one graph convolution layer. A fully connected layer with softmax loss is usually used after graph convolution layers for classification.

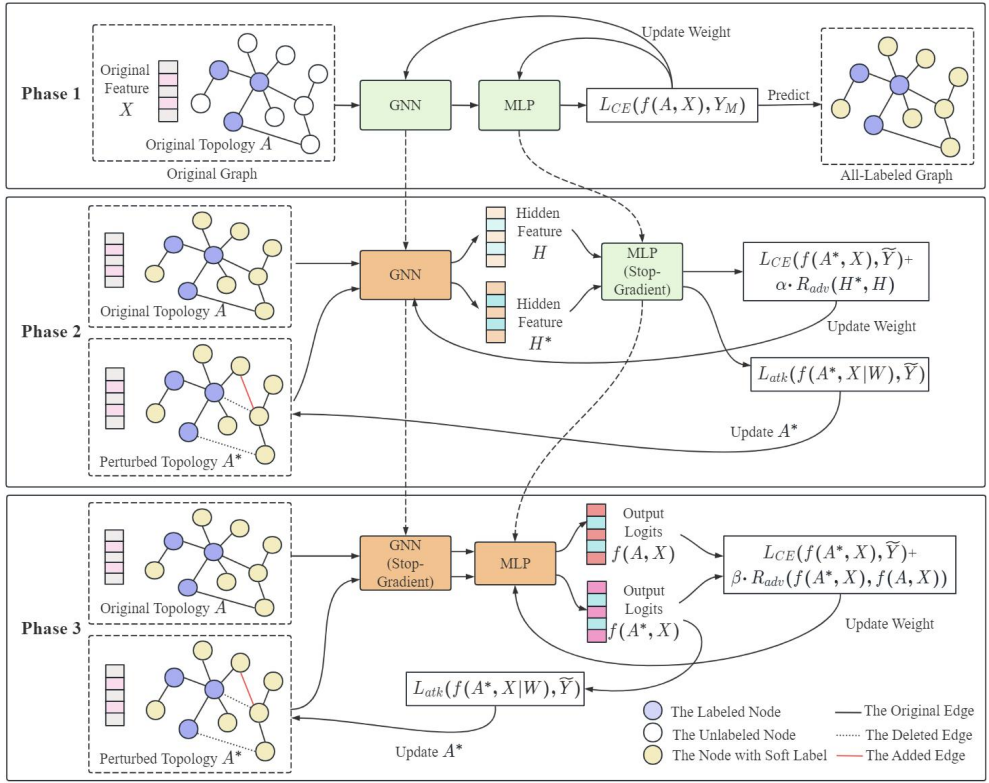

**Fig. 1.** The overall framework of HC-Ref.

*B. Adversarial Training on Graphs*

Adversarial training is an important way to enhance the robustness of neural networks. In adversarial training, the samples are mixed with some small perturbations (the changes are small, but it is likely to cause misclassification), and then the model is prompted to adapt to the change, thus being robust to adversarial samples. On the graph, they can be divided into perturbation of node features and perturbation of graph structure.

*1) Adversarial training based on node features*: The perturbation of node features can be defined as follows:

$$r^* = \underset{r}{argmax} \sum_{i \in V} L(y, \hat{f}(x + r, \bar{x}, A|W)) \quad (2)$$
$$s.t. \|r^*\| < \epsilon$$

where $r^*$ represents the perturbation of a specific node, $\bar{x}$ represents features excluding feature $x$ of current node, $\epsilon$ denotes the limit of perturbation and $W$ is the parameters of the model trained in the previous phase.

Then, to, the goal of adversarial training can be described as follows:

$$W^* = \underset{W}{argmin} \sum_{i \in V} L(y, \hat{f}(x + r^*, \bar{x}, A|W)) \quad (3)$$

*2) Adversarial training based on graph structure*: Similar to (2), the perturbation of graph structure is obtained by the following equation:

$$A^* = \underset{A}{argmax} \sum L(Y, \hat{f}(X, A|W)) \quad (4)$$
$$s.t. \frac{\|A^* - A\|_0}{\|A\|_0} < \epsilon$$

Different from continuous changes in features, the perturbation of graph structure is discrete. Therefore, some methods search for $A^*$ through randomization [4] or greedy strategies [5], [28]. Reinforcement learning is also used to generate $A^*$ [26]. Noting that for the first time, Xu *et al.* [29] constructs $A^*$ from an optimized perspective.

Then, the goal of adversarial training with perturbation of graph structure can be described as (5):

$$W^* = \underset{W}{argmin} \sum L(Y, \hat{f}(X, A^*|W)) \quad (5)$$

Besides, in some work [15], [29], (4) and (5) are used iteratively to obtain a more robust model.

III. PROPOSED APPROACH

In this section, we will first design adversarial samples based on topological perturbations and a universal adversarial training pattern. Then, a smooth regularization term for graph-based adversarial training is introduced, followed by refinement strategies for feature extractors and downstream classifiers, respectively. Finally, the overall framework of HC-Ref is presented, which is shown in **Fig. 1**.

### A. Adversarial Training on Graphs

*1) Adversarial samples based on topological perturbations*: Before conducting adversarial training on graphs, it is necessary to choose an appropriate way to generate adversarial samples. A common approach [4] is to construct adversarial samples based on the gradient of the loss function, which can be used to most effectively simulate perturbations in the worst-case scenario. In contrast, although methods based on greedy strategies [5], [28] can find local optima at each step during the search process, they cannot guarantee that the generated adversarial samples are better globally. Methods based on reinforcement learning [26] inevitably introduce additional time costs. Optimization-based methods iteratively use the local first-order gradients during model training, and gradually approach the optimal perturbation based on the principle of local linearity. Thus, we recommend generating adversarial samples from an optimization perspective. However, although it is feasible to apply similar concepts to transfer the extreme perturbation onto changes in the graph structure, there are two challenges that must be addressed: 1) Graph structures are discrete, while gradient-based attacks are continuous, so a specialized scheme needs to be designed to bridge the gap between them; 2) Training the GNN model solely using the information of labeled nodes makes it difficult for the unlabeled high-order neighbors of training nodes to benefit from adversarial training.

To solve the first problem, inspired by Xu *et al.* [29], we extend the discrete elements in the adjacency matrix $A_{ij} \in \{0,1\}$ to a continuous space $A_{ij} \in [0,1]$ through convex relaxation. To obtain the topological change matrix $S$, an attack loss $L_{atk} \in \{L_{CE}, L_{CW}\}$ similar to the Cross-Entropy Loss (CE-loss) [33] or Carlili-Wagner Loss (CW-loss) [34] is used, which can be described as:

$$L_{CE} = -\sum_{j \in V_l} \sum_{i=1}^{C} y_i \log \hat{y}_i \quad (6)$$

$$L_{CW} = \sum_{i \in V_l} max\{Z_{i,y_i} - max\{Z_{i,c}; c \neq y_i\}, -k\} \quad (7)$$

where $V_l$ represents the set of labeled nodes, $C$ denotes the number of categories, $Z_{i,c}$ represents the logical value corresponding to the category $c$ of the node with index $i$ and $k$ is a hyperparameter.

We generate topological perturbations by iteratively performing projected gradient descent, as shown below:

$$s^{(t)} = \prod_S(s^{(t-1)} - \lambda \mu^{(t)} \nabla L(s^{(t-1)})) \quad (8)$$

where $t$ denotes the iteration index of PGD, $\lambda > 0$ is the learning rate of attack, $\mu^{(t)}$ is the adaptive factor during attack, $\nabla L(s^{(t-1)})$ denotes the gradient of the attack loss $L$ evaluated at $s^{(t-1)}$, $\prod_S(a) := argmin_{s \in S} \|s - a\|_2^2$ is the projection operation, as shown in (9), which solves the constraint problem in formula (4): $\|A^* - A\| < \epsilon$. Finally, the elements in the topological change matrix S can be translated into edge perturbation probabilities, and random sampling [35] will be applied to obtain binary topological perturbation $\hat{S}$.

$$\prod_S(a) = \begin{cases} P_{[0,1]}[a], \text{if } \text{sum}(P_{[0,1]}[a]) \leq \epsilon \\ P_{[0,1]}[a - \mu], \substack{\text{if } \mu > 0 \text{ and} \\ \text{sum}(P_{[0,1]}[a-\mu]) \leq \epsilon} \end{cases} \quad (9)$$

where sum represents the sum of all elements and $P_{[0,1]}(x) = x$ if $x \in [0,1]$, 0 if $x < 0$ and 1 if $x > 0$.

Finally, $A^*$ is obtained by the following equation:

$$A^* = A + (\bar{A} - A) \circ \hat{S} \quad (10)$$

where $\hat{S} \in \{0,1\}^{N \times N}$ is a binary topological perturbation obtained from section 3.1. $s_{ij} \in \hat{S}$ represents whether the edge between vertex $v_i$ and vertex $v_j$ has been perturbed. If $s_{ij} = 1$, then delete the edge between vertex $v_i$ and vertex $v_j$ or add a new edge between them if there was no edge before. ∘ represents the Hadamard product. Given an adjacency matrix $A$, its complement matrix is $\bar{A} = E - I - A$. $E$ is a matrix with all elements being 1 and $I$ is an identity matrix.

To solve the second problem, we adopted the Robust Self-Training (RST) strategy. Firstly, a clean GNN model is trained by using the features of all nodes, the adjacency matrix, and a small amount of obtainable labels. It then performs the inference task to provide soft labels for unlabeled nodes. That is:

$$\tilde{y}_i = \begin{cases} y_i, & i \leq M, \\ \hat{f}(X, A|W), & M < i \leq N \end{cases} \quad (11)$$

where $M$ represents the number of all labeled nodes.

By combining the true labels of trained nodes and the soft labels of remaining nodes, the model can perform supervised learning, so that all nodes will benefit from adversarial training, thus avoiding the second problem mentioned above. RST helps to improve the robustness and generalization of the network, which has been demonstrated both theoretically and experimentally by Carmon et al. [36].

*2) Adversarial Training*: This paper focuses on topology-based robustness training. Under this setting, both the original features of the nodes and the perturbed graph topology will be used for model training.

Specifically, we use iterative forms of (4) and (5), which are (12) and (13), to conduct adversarial perturbation and adaptive training correspondingly, since local optimization helps to gradually approach the target where the model is more robust.

$$A^*_{t+1} = \underset{A}{argmax} \sum L(Y, \hat{f}(X, A^*_t | W_t)) \quad (12)$$

$$s.t. \frac{\|A^*_{t+1} - A\|_0}{\|A\|_0} < \epsilon$$

$$W_{t+1} = \underset{W}{argmin} \sum L(Y, \hat{f}(X, A^*_{t+1} | W_t)) \quad (13)$$

*B. Smoothness Regularization on Graphs*

*1) Motivations*: Some works [37], [38] have theoretically proven that in adversarial training, if the optimization objective simply involves minimizing the cross-entropy between the predicted and correct labels of adversarial samples, it is easy to encounter an unstable situation: The probability, $\Pr[X \in \mathbb{B}(DB(f), \epsilon), f(X)Y > 0]$, is high that the data is located near the decision boundary of the model, indicating that there is a risk that small perturbations may move data points to the incorrect side of the decision boundary. As a result, there is a contradiction between the original design of the loss function and training a more robust model. In other words, there is a great deviation between the reached point of robustness gain and the model space with more robustness. Furthermore, this problem is even more significant for GCNs trained on perturbed topologies, because the presence or absence of an edge $e_{ij}$ directly affects whether the information $x_i$ of a node $v_i$ can be transferred to its neighbor $v_j$. Repeatedly adjusting the adjacency matrix $A^*$ will cause significant changes in the output of GCN $H^{(t)}$, which directly increases the difficulty of the optimization task (slow convergence speed) and causes fluctuations in the performance of downstream task (with a large variance in statistical data). In addition, the aforementioned deviation problem still exists. To address these challenges, our first work attempts to add a smooth regularization term in graph adversarial training. On the one hand, by introducing a prior distribution to guide and reduce the model's sensitivity to these changes, it reduces learning difficulty and mitigates fluctuations by tightening upper and lower bounds; on the other hand, it establishes smoother and more robust decision boundaries.

*2) Method Formulation.* Inspired by Trades [38], the smoothness helps the model to improve its robustness and tighten the upper and lower bounds of the objective loss by reducing the prediction difference between natural samples and adversarial samples. The graph smoothness regularization we proposed can be described as:

$$R_{adv}(Z^*, Z) = D(f_s(Z^*), f_s(Z)) \quad (14)$$

where $f_s(Z) = \{\frac{e^{z_i}}{\sum_{c=1}^C e^{z_c}} | i = 1,2,\dots,C\}$ converts $Z$ into a probability distribution, $D$ represents the distance between two distributions, and in this paper we use KL divergence as an alternative. We use the regularization at different stages.

The purpose of optimizing the reduction of this term is to encourage the smoothness of the model output, in other words, it drives the decision boundary of the model away from sample instances by minimizing the difference between the prediction of natural parts $f(X, A|W)$ and adversarial parts $f(X, A^*|W)$. This is conceptually consistent with the argument [39] that smoothness is an indispensable property of robust models. When updating the network through optimizer during adversarial training using the traditional CE-Loss, it inevitably leads to sharp decision boundaries (in order to include some distant adversarial samples exactly in the corresponding category). By introducing an inductive bias and forcibly smoothing these sharp boundaries, the overfitting of the network is alleviated.

*C. Hierarchical Refinement*

*1) Motivations*: We noticed that some studies [31], [32] focus on training certain layers in the network and improve the inference capability of the model, which inspires us that some special knowledge extraction strategies have the potential to further improve robustness. Similarly, our second work proposes a hierarchical refinement training strategy. Here, we utilize GCN to be a surrogate model for GNN [1], which is used as a feature extractor that maps node features into high-level semantic features. And MLP with only a single fully connected layer will be used as a simplified surrogate model for classifiers that perform downstream tasks. Hierarchical refinement means designing corresponding adversarial regularization terms for different layers of the model and independently training them in different phases. In Section IV, we carefully designed a series of experiments to demonstrate the superior performance of hierarchical refinement and attempted to explain the reason for the effectiveness of our method.

*2) Standardized training*: A common phenomenon in training GNNs is that the cross-entropy loss rapidly decreases during its initial training phase. In other words, the model weights are iterating to the low-loss regions, and far away from falling into the sharp local minima in the loss landscape. However, adversarial perturbations can affect the convergence speed in the initial phase and even mislead GNNs towards worse model spaces [40]. This problem encourages us to standardize the training of GNNs in the first phase, and in the later phases, regularize the GNNs through adversarial training. It is more rational to reach the low-loss regions through standard training, avoiding the aforementioned problem. At this point, adversarial training tends to fall into local minima that make model weights more robust. And the objective function of phase 1 is shown below:

$$L_{p_1} = L_{CE}(\hat{f}(A, X), Y_M) \quad (15)$$

where $Y_M$ represents the labels of all labeled nodes.

*3) Refining feature extractor*: In this phase, we focus on the robust learning of GNN, which is regarded as feature extractor. First, the model parameters obtained in the first phase will be further trained. Some works [31], [50] suggest freezing the classifier and using the boundary of the natural classifier as a guide for training in adaptive training to achieve better results. Such an approach reduces the difficulty of training and helps to learn cleaner features. We will also freeze the downstream classifier trained with standardization in this stage, in other words, the parameters of the MLP are fixed. Intuitively, the optimization approach causes the standardized classifier to favor high-level semantic features that are more similar to those generated by natural samples, forcing the GNN to extract high-level semantic features from adversarial samples that are similar to those of natural samples. However, optimization itself is very difficult and biased because adversarial samples are generated by worst-case perturbations, making it difficult to produce representations close to the natural samples. A appropriate loss function is needed to solve the problem.

The solution proposed by Robust GCN [41] is to force the features of each layer to follow a standard Gaussian distribution. However, even if the features are constrained to the known domain in this way, avoiding out-of-distribution (OOD) issues, it greatly limits the expressive power of the model. We propose to constrain the high-level semantic features of adversarial samples by the following formula:

$$R_{adv}(H^*, H) = -\sum f_s(H^*) \log \frac{f_s(H)}{f_s(H^*)} \quad (16)$$

where $H^*$ and $H$ are high-level semantic features extracted from adversarial samples and natural samples, respectively. $f_s(H) = \frac{e^{H_i}}{\sum_{c=1}^{C} e^{H_c}}$ converts high-level semantic features $H$ into probability distributions.

Equation (16) shortens the domain gap between the high-level semantic feature distributions of the two sample groups and promotes smoothness between them, thus enhancing the robustness of the model [39]. Noting that it is not a standard distance metric because it is asymmetric. But it is rational because our goal is to correct the model's representation of adversarial samples, rather than constraining that of the natural samples. We believe that this (constraining the prediction of natural samples) changes the way information is transmitted, which is not conducive to guiding training. The final optimization process is shown in phase 2, as illustrated in Figure 1. And the objective function of phase 2 is shown below:

$$L_{p_2} = L_{CE}(f(A^*, X), \widetilde{Y}) + \alpha \cdot R_{adv}(H^*, H) \quad (17)$$

where $\alpha$ is a trading hyper-parameter.

4) *Refining downstream classifiers*: In this phase, the classifier that performs the downstream task will be fine-tuned meticulously. The feature extractor trained in phase 2 now continuously provides the classifier with purer and easier-to-learn features. As a result, here we only need to focus on enhancing the classification ability of the MLP. Similar to phase 2, the GNN serving as a feature extractor is frozen in phase 3, which is necessary in that further training of an unfrozen GNN can lead to suboptimal results due to overfitting, as confirmed during subsequent experiments. The final optimization process is shown in phase 3, as illustrated in Figure 1. And the objective function of phase 3 is shown below:

$$L_{p_3} = L_{CE}(f(A^*, X), \widetilde{Y}) + \beta \cdot R_{adv}(f(A^*, X), f(A, X)) \quad (18)$$

where $\beta$ is a tunable hyperparameter that represents the trade-off between $L_{CE}$ and $R_{adv}$.

The overall training process is shown in Algorithm 1.

---

**Algorithm 1 The HC-Ref algorithm**
**Input**: The number of training steps $T_{train}$. The perturbation rate $\epsilon$. The attack iterations $T_{atk}$. Learning rate $\eta_t$. The adjacency matrix $A$. The feature matrix of nodes $X$.

1: Randomly initialize the model parameters of phase 1: $W_{p_1}^0 = W^0$
2: **for** $t = 1,2 \ldots T_{train}$ **do**
3:    Update the model parameters: $W_{p_1}^t = W_{p_1}^{t-1} + \eta_t \nabla L(f(X, A|W_{p_1}^{t-1}), Y)$
4: **end for**
5: Set the model parameters of phase 2: $W_{p_2}^0 = W_{p_1}^{T_{train}}$
6: **for** $t = 1,2 \ldots T_{train}$ **do**
7:    Set $S^{(0)} = 0$;
8:    **for** $i = 1,2 \ldots T_{atk}$ **do**
9:       get $S^{(i)}$ by (8) and projection operation (9);
10:    **end for**
11:    obtain the binary topology perturbation matrix $\hat{S}$ by random sampling from probabilistic topology perturbation $S^{(T_{atk})}$;
12:    get $A^*$ by (10);
13:    Update the model parameters: $W_{p_2}^t = W_{p_2}^{t-1} + \eta_t \nabla L_{p_2}$, where $L_{p_2}$ is showed in (17) and the parameters of MLP are fixed;
14: **end for**
15: Set the model parameters of phase 3: $W_{p_3}^0 = W_{p_2}^{T_{train}}$
16: **for** $t = 1,2 \ldots T_{train}$ **do**
17:    repeat step 7- step 12;
18:    Update the model parameters: $W_{p_3}^t = W_{p_3}^{t-1} + \eta_t \nabla L_{p_3}$, where $L_{p_3}$ is showed in (18) and the parameters of GNN are fixed;
19: **end for**

---

## IV. EXPERIMENT

In this section, our goal is to verify the superior robustness of our method by performing a node classification task (node classification is the most common task, aimed at determining the label of vertices.). In the following experiments, it can be observed that the model trained by HC-Ref can still maintain a high classification accuracy under attack and perform better than various advanced methods.

*A. Experimental Setup*

*1) Datasets*: We conduct experiments on two types of node classification datasets: Cora [52] and Citeseer [52], two benchmark citation networks. Nodes represent documents, and edges represent reference links between documents. Each node has a feature vector (which is a sparse bag representa-tion of the document) and an artificially annotated topic as a class label. The vertices of Cora fall into 7 classes and that of Citeseer are divided into 6 categories. The statistics of the experiment datasets are summarized in Table II. In training, we use the features of all nodes and the labels of the top M (The number of labeled node, which is 140 for Cora and 120 for Citeseer) nodes. The 500 nodes after the nodes in the training set are used for validation, and the remaining nodes for testing.

TABLE III
NODE CLASSIFICATION RESULTS ON CORA ($\epsilon$ DENOTES PERTURBATION RATES)

| | | CE-PGD | | | | CW-PGD | | | |
|---|---|---|---|---|---|---|---|---|---|
| $\epsilon$ | | 5% | 10% | 15% | 20% | 5% | 10% | 15% | 20% |
| Methods | GCN | 0.755 | 0.719 | 0.699 | 0.670 | 0.737 | 0.657 | 0.600 | 0.561 |
| | Random | 0.762 | 0.720 | 0.684 | 0.649 | 0.724 | 0.655 | 0.588 | 0.559 |
| | GraphAT | 0.783 | 0.729 | 0.712 | 0.692 | 0.747 | 0.704 | 0.665 | 0.604 |
| | TGD | 0.784 | 0.773 | 0.751 | 0.739 | 0.779 | 0.771 | 0.748 | 0.721 |
| | HC-Ref | **0.803** | **0.789** | **0.773** | **0.748** | **0.782** | **0.776** | **0.754** | **0.747** |

TABLE IV
NODE CLASSIFICATION RESULTS ON CITESEER ($\epsilon$ DENOTES PERTURBATION RATES)

| | | CE-PGD | | | | CW-PGD | | | |
|---|---|---|---|---|---|---|---|---|---|
| $\epsilon$ | | 5% | 10% | 15% | 20% | 5% | 10% | 15% | 20% |
| Methods | GCN | 0.636 | 0.601 | 0.578 | 0.528 | 0.630 | 0.597 | 0.564 | 0.509 |
| | Random | 0.633 | 0.596 | 0.566 | 0.539 | 0.625 | 0.594 | 0.548 | 0.512 |
| | GraphAT | **0.696** | 0.673 | 0.643 | 0.614 | **0.682** | 0.665 | 0.623 | 0.597 |
| | TGD | 0.684 | 0.664 | 0.643 | 0.615 | 0.673 | 0.661 | 0.646 | 0.627 |
| | HC-Ref | 0.689 | **0.678** | **0.661** | **0.646** | 0.677 | **0.669** | **0.654** | **0.639** |

TABLE II
STATISTICS OF THE EXPERIMENT DATASETS

| Dataset | Nodes | Edges | Classes | Features | Label rate |
|---|---|---|---|---|---|
| Cora | 2,708 | 5,429 | 7 | 1,433 | 0.052 |
| Citeseer | 3,312 | 4,732 | 6 | 3,703 | 0.036 |

*2) Baselines*: We compared our approach to the following: 1)Random [4]. It randomly removes edges (within budget) at each training stage. 2) GraphAT [25]. It optimizes additional regularization terms to reduce the impact of perturbation propagation through node connections. Furthermore, it introduces Virtual Adversarial Training [24], which constructs adversarial samples based on the local distributional smoothness of unlabeled nodes. 3) TGD [29]. It updates model parameters from an optimized perspective by iteratively attacking and defending. We use its retrained version rather than the fixed version because the former has better performance on defense [29].

*3) Implementation Details*: When conducting the node classification task, we use a two-layer GCN followed by an MLP as the basic framework, as mentioned in Section III. Here, we provide detailed implementation details of our experiments.

For GCN, we set the dimension of hidden features h=32 for each layer. For MLP, we only use a single fully connected layer and use Relu for final activation, which can be represen-ted as $z = \sigma(wx + b)$. Then, we set learning rate = 0.01, drop out = 0.0, weight deacy = 5e-4 and the training epoch is 120. In the process of generating adversarial samples, we set the number of attack iterations per epoch of training $T_{atk}$= 40, the learning rate of attack $\lambda$=1, and the perturbation rate $\epsilon \in \{0.05, 0.1, 0.15, 0.2\}$. To avoid fluctuation in the later stages of model learning, we add an adaptive factor $\mu^{(t)} = \frac{\mu^{(0)}}{(t+1)^2}$. To be specific, we set $\mu^{(0)}$=200 in CE-PGD, and $\mu^{(0)}$=0.1 in CW-PGD. In Cora, $\alpha$ is 16 (trade-off value in $L_{p_2}$), $\beta$ is 32 (trade-off value in $L_{p_3}$), and in Citeseer, both are 0.05. The parameter sensitivity analysis of $\alpha$ and $\beta$ has been shown in Section IV.

In Random, the low drop rate results in little change in the topology, making it difficult to produce gains. On the other hand, a drop rate that is too high severely disrupts the original structure, leading to poor perfomance. Therefore, we set the proportion of edges to be dropped in each round moderately, that is, del_rate $\in [0.05, 0.2]$. Other parameters remain the same as [4], and the best results are recorded.

In GraphAT, we follow the suggested settings in [25], using virtual adversarial loss and graph adversarial loss, and set corresponding weights alpha $\in [0.1, 1.0]$ and beta $\in [0.1, 1.0]$. Other parameters are also kept consistent with [25], and only the best results are recorded.

In TGD, we set the perturbation rate $\epsilon \in \{0.05, 0.1, 0.15, 0.2\}$. To ensure fairness in the comparison, the above methods were unified with the same network structure and general training parameters as ours.

Another potentially controversial point is that HC-Ref used pre-training for model training. Thus, we set the same pre-training process for other compared models as well.

*B. Comparison Results*

*1) Defensive performance of different models*: As shown in Table III and Table IV, we compared the performance of each

baseline with our method under two attacks, CE-PGD [29] and CW-PGD, with 100 attack iterations for each. GCN is considered as a control group without using any defense mechanisms. In addition, we highlight the best results in bold and underline the second best ones. Observing the results, we obtained the following conclusions:

1) Random only shows improved robustness in certain cases. For instance, it displays some level of resilience against CE-PGD attacks (with low perturbation rates) on Cora. However, in most other cases, its robustness is even worse than that of the original GCN.

2) GraphAT demonstrates superior performance to GCN under various attacks with different perturbation rates, indicating that the approach possesses stable defense capabilities against such attacks. However, the model's adversarial robustness is only strong at lower perturbation rates (<10%). With an increase in the perturbation rate, the accuracy of GraphAT will decrease and reach a lower level.

3) TGD has stable and better defense capabilities, which can be maintained even at higher perturbation rates. However, it can be further improved under our framework.

4) In comparison to the aforementioned methods, HC-Ref typically demonstrates superior predictive performance under various attacks with different perturbation rates. Moreover, it exhibits more significant robustness when confronting attacks with higher perturbation rates.

TABLE V
THE SUCCESS RATE OF THE ATTACKS

| $\epsilon = 5\%$ | Cora | | Citeseer | |
| --- | --- | --- | --- | --- |
| | GCN | HC-Ref | GCN | HC-Ref |
| Random | 1.5% | 0.4% | 2.6% | 0.6% |
| DICE | 1.8% | 0.7% | 3.8% | 1.3% |
| Meta-Self | 5.6% | 1.8% | 5.2% | 2.9% |
| CE-PGD | 5.9% | 1.1% | 5.5% | 0.2% |
| CW-PGD | 7.7% | 3.2% | 6.1% | 1.4% |
| AtkSE | 8% | 3.4% | 10.4% | 3.7% |
| GraD | 10.3% | 4.1% | 14.1% | 4.5% |

2) *Defensive performance under different attacks*: To validate whether HC-Ref has the capacity to generalize to defend against other attack strategies, we evaluated the performance of HC-Ref under various attack methods as well. These attack methods include Random and DICE [47] based on random sampling, as well as AtkSE [48], GraD [49] and TGD [29] based on gradients. The specific results are shown in Table V. To ensure fairness of the results, the attack budget $\epsilon$ was set uniformly to 5%. The parameters of different attack methods were set according to their default optimal settings in the original papers. Notice that here we present the success rate of the attack $\omega = (N_{w_{att}} - N_{w_{nat}})/N$ as the evaluation metric. $N_{w_{att}}$ represents the number of misclassified samples after the attack, $N_{w_{nat}}$ represents the number of misclassified samples before the attack, and we consider the difference between them as the impact generated by the attack. Obviously, our method significantly improves defense capability of the network against various (random-based or gradient-based) attack strategies.

3) *Performance on the original graph*: Furthermore, we discuss the generalization ability of HC-Ref. The evaluation on the original graph is shown in Table VI. For simplicity, the other methods in comparison only show the optimal results under various hyperparameters. The results indicate that our method not only exhibits superior performance in defending against attacks but also provides better generalization capabilities for GNNs. The phenomenon indicates that improving the robust accuracy on discrete graph topology does not require a reduction in natural accuracy. HC-Ref achieves the highest accuracy of 0.824 on Cora and the second-highest accuracy of 0.706 on Citeseer. In this case, the most competitive method is GraphAT. It is worth noting that HC-Ref exhibits the best performance on different datasets with different settings (a perturbation rate of 0.2 when trained on Cora, and a perturbation rate of 0.05 when trained on Citeseer, both using adversarial examples generated by CE-PGD for training). Therefore, the choice of hyperparameters is critical, which is analyzed in detail below.

TABLE VI
NODE CLASSIFICATION RESULTS ON ORIGINAL GRAPH

| Method | Cora | Citeseer |
| --- | --- | --- |
| GCN | 0.814 | 0.691 |
| Random | 0.800 | 0.673 |
| GraphAT | 0.820 | **0.708** |
| TGD | 0.815 | 0.692 |
| HC-Ref ($\epsilon = 5\%$) | 0.816 | 0.706 |
| HC-Ref ($\epsilon = 10\%$) | 0.818 | 0.692 |
| HC-Ref ($\epsilon = 15\%$) | 0.819 | 0.695 |
| HC-Ref ($\epsilon = 20\%$) | **0.824** | 0.699 |

*C. Ablation Study*

In this section, our goal is to ensure that each step in our method can deliver practical advantages.

As an instance, we conducted ablation experiments on Cora using the CE-PGD attack, perturbation rate of 0.2 ($\alpha=0.05$, $\beta=0.05$) to highlight the changes in evaluation more prominently. For the sake of simplicity, we will use the following notations for replacement: HC-Ref is our method. HC-1 trains the feature extractor through (14) and freezes the classifier. HC-2 trains the classifier through (15) and freezes the feature extractor. Cons-H trains the complete network through (14). Cons-D trains the complete network through (15). HC-Uncon follows the same process as HC-Ref, except that no freezing strategy is adopted in both stages. The design is intended to answer the following questions:

1) Would refining the GNN or classifier separately be effective?

2) Would utilizing regularization separately to restrict the distance between hidden features or predicted distributions result in improvement, and if so, to what degree?

3) What impacts will be produced by training without freezing strategy?

**Fig. 2** shows the node classification accuracy evaluated after conducting attacks on each control group as the number of epochs increases. To present the results more clearly, the horizontal lines in the figure indicate the mean node classification accuracy from epoch 120 to 240, with different colors corresponding to different experimental groups. The following conclusions can be summarized:

1) Among all the experimental groups, HC-Ref achieved the best performance in terms of node classification accuracy under attack. Specifically, after being attacked, HC-Ref exhibited a more compact upper and lower bound and a higher mean accuracy compared to the other experimental groups, which means that the combination of hierarchical constraints (different task goals for the model at different stages) and refinement strategies (freezing certain layers) is more effective than other combinations in improving the robustness of the model to adversarial attacks.

2) The effectiveness of refining only the GNN (HC-1) also confirms our analysis of the benefits of refinement for the feature extractor, as discussed in Section III. In contrast, refining only the classifier (HC-2) does not improve the robustness of the GCN model, and even leads to worse performance compared to the original GCN model. As a matter of fact, the feature extractor was frozen without being robustly trained, leading to the extraction of fragile features that are highly vulnerable to even slight adversarial attacks. The fragility of these hidden features may have contributed to the inability of a single fully connected layer to effectively defend against attacks based on the gradient of the entire model.

3) Constraining only the hidden features (Cons-H), constraining only the output layer logits (Cons-D, similar to Trades), and training without freezing strategy (HC-Uncon) can also improve robustness. Overall, the different groups showed comparable performance in terms of node classification accuracy, as indicated by the overlap of the lines representing their performance. However, these methods cannot achieve the same level of improvement as HC-Ref.

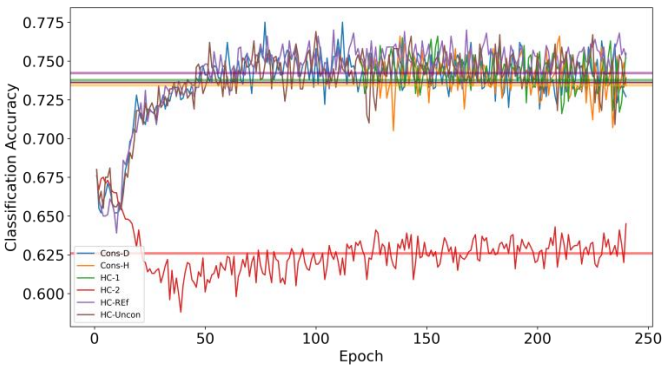

**Fig. 2**. Ablation Experiment: The node classification accuracy of HC-Ref and the other five experimental groups under adversarial training after each epoch, as well as the average accuracy (horizontal lines) over the last 120 epochs.

*D. Parameter Analysis*

In this section, we adjusted and analyzed the hyper-parameters in HC-Ref method. The hyperparameters that have the greatest impact on the model prediction are: The perturbation rate $\epsilon$, the number of attack iterations $T_{atk}$, learning rate $\eta_t$, trade-off value $\alpha$ and $\beta$.

TABLE VII
MISCLASSIFICATION RATES OF CE-PGD ATTACK AGAINST ROBUST TRAINING MODEL UNDER DIFFERENT $\epsilon$.

| Misclassification rates | | $\epsilon$ in robust training (in %) | | | | |
|---|---|---|---|---|---|---|
| | | 0 | 5 | 10 | 15 | 20 |
| $\epsilon$ in attack (in %) | 0 | 18.6 | 18.4 | 18.2 | 18.1 | 17.6 |
| | 5 | 24.5 | 19.3 | 19.7 | 20.8 | 21.3 |
| | 10 | 28.1 | 22.5 | 21.1 | 22.0 | 22.9 |
| | 15 | 30.1 | 25.3 | 22.7 | 22.3 | 23.4 |
| | 20 | 33.0 | 27.8 | 25.2 | 24.1 | 23.8 |
| | average | 26.9 | 22.7 | **21.4** | 21.5 | 21.8 |

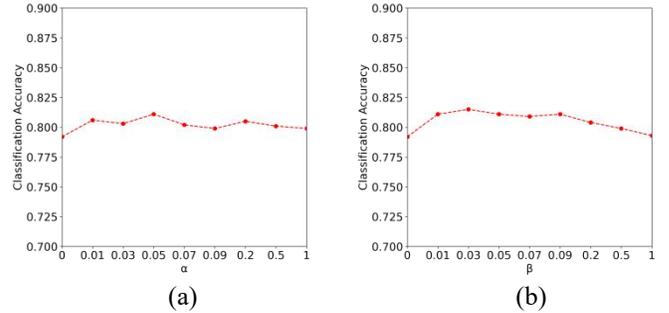

**Fig. 3**. Analysis of Parameter $\alpha$ (a) and $\beta$ (b)

We aim to show which $\epsilon$ we need to use in robust training, so we set $\epsilon \in \{0.05, 0.1, 0.15, 0.2\}$ and apply CE-PGD attack following the same setting. The results on Cora are presented in Table VII. We can draw the conclusion that under attacks with different levels of perturbation, models trained with corresponding perturbation rates have lower misclassification rates. Considering the results on the original graph and various perturbed graphs, we suggest setting $\epsilon$ to 0.1 during training. $T_{atk}$ is the number of attacks before each round of model parameter update. Too many attacks will lead to unnecessary time consumption, and too few attacks will make it difficult to ensure the quality of the generated adversarial samples. Therefore, we set $T_{atk}=40$. In this section, we set the learning rate $\eta_t$ to 0.01. A bigger learning rate may lead to divergence, while a smaller learning rate may lead to slow convergence. Then, we analyzed the impact of different values of $\alpha$ and $\beta$ on the model performance, and the results are shown in **Fig. 3**. We set $\alpha=0.05$ when analyzing the impact of changes in $\beta$, and $\beta=0.05$ when analyzing the impact of changes in $\alpha$. It is evident that they have similar trends. Within a certain range, increasing the parameter $\alpha$ (or $\beta$) was found to improve the performance of the model by providing greater smoothness to the optimization process. It is worth noting that if the parameter $\alpha$ (or $\beta$) exceeds this range, the impact of the regularization becomes significantly greater than that of the CE-Loss, which in turn leads to a decrease in performance. In such cases, the optimization objective is no longer dominated by classification accuracy, but rather by smoothness, which leads to excessive similarity between the natural and adversarial components, and is not conducive to classifying accurately.

## V. Related Work

*A. Adversarial Defense on Graphs*

The objective of adversarial defense on graphs is to preserve a high level of predictive performance even when attackers introduce deliberately crafted perturbations to the graph. The primary approaches in adversarial defense on graphs can be broadly classified into three categories: self-supervised-based method, detection-based method, and adversarial-based method. The self-supervised-based method [42]–[44] utilizes self-information to enhance the robustness of GCNs. The detection-based method [14], [30] aims to mitigate the negative impact of attacks by detecting and removing potential attacker nodes or edges. In this paper, we mainly discuss the methods based on adversarial training [26], [29], [45], which learn a more robust model and resist adversarial attacks by training with adversarial examples. The difficulty of the work lies in finding typical adversarial examples that can be incorporated into the training to provide robustness gains. Furthermore, our approach takes into account carefully crafted training strategies that can further enhance performance.

*B. Adversarial Training with GNNs*

Adversarial training can effectively improve the robustness of GNNs [15]. Some works [21], [22], [25] introduce virtual adversarial training [24] on the graph for adversarial training and generate adversarial examples based on local distributional smoothness (LDS). Kong *et. al.* [46] adopted the concept of FreeAT [17] into GNN adversarial training and utilized neighbor sampling strategies to expand the approach to super large-scale graphs. However, we noticed that many GNN adversarial training works solely add perturbations into node features, disregarding the significant impact of graph structure.

The first adversarial training study that employed graph structure perturbations [4] randomly removed edges during each training iteration. However, the outcomes revealed that this approach had minimal impact on improving robustness. Several studies [30], [51] leverage the concept of modifying graph topology and adversarial training to improve the generalization capability of GNNs for out-of-distribution scenarios. Nonetheless, there is no proof indicating that these approaches can effectively defend against adversarial attacks. The work that is closest to ours is TGD [29], which reduces the sensitivity of the model to graph structure perturbations by training GNN on perturbed topology. However, we regret to discover that the above research solely performs adversarial training on the perturbed graph topology, neglecting the knowledge that could be acquired from the original, unperturbed graph topology, which is further studied in this paper. Furthermore, a carefully designed hierarchical training framework is used to refine different layers of the model, significantly enhancing adversarial robustness without changing the network.

## VI. Conclusion

HC-Ref is a novel method for graph adversarial defense that accepts guidance from regularized constraints to improve smoothness under the framework of adversarial training, which also emphasizes independent training during different phases of the model. The experiments show that HC-Ref can effectively defend against graph-based attacks, improve adversarial robustness, enhance the model's generalization ability on the original graph, and result in better performance for downstream tasks. We further conducted extensive experiments on our approach and obtained intriguing and valuable results and explanations.

One of our future directions is to extend HC-Ref to more downstream tasks, such as link prediction, graph classification, and applications such as recommendation systems, fraud detection, etc. Another interesting direction is to extend our method to directed and weighted graphs, and further investigate its ability to generalize to dynamic or heterogeneous graphs. Furthermore, it would be beneficial to develop approximation methods to expedite the adversarial training process or subgraph sampling strategies to apply our approach to large-scale graphs. Careful design of these techniques could significantly improve the scalability and efficiency of our method.